\pdfoutput=1

\documentclass[11pt]{article}

\usepackage{acl}

\usepackage{times}
\usepackage{latexsym}

\usepackage[T1]{fontenc}

\usepackage[utf8]{inputenc}

\usepackage{microtype}

\usepackage{amsmath}
\usepackage{amssymb}
\usepackage{graphicx}
\usepackage{booktabs}
\usepackage{multirow}
\usepackage[labelfont={bf}]{caption}
\usepackage[usestackEOL]{stackengine}
\usepackage{enumitem}

\newcommand\blfootnote[1]{%
  \begingroup
  \renewcommand\thefootnote{}\footnote{#1}%
  \addtocounter{footnote}{-1}%
  \endgroup
}
\newcommand\Tstrut{\rule{0pt}{2.2ex}}         % = `top' strut
\newcommand\Bstrut{\rule[-0.9ex]{0pt}{0pt}}   % = `bottom' strut

\title{Privacy-Preserving Text Classification on BERT Embeddings \\
with Homomorphic Encryption}

\author{Garam Lee\textsuperscript{*1} \quad
  Minsoo Kim\textsuperscript{*2} \quad
  Jai Hyun Park\textsuperscript{*2} \\
  {\bf Seung-won Hwang\textsuperscript{\textdagger2} \quad
  Jung Hee Cheon\textsuperscript{1,2}} \\
  \textsuperscript{1}{CryptoLab} \quad \textsuperscript{2}{Seoul National University} \\
  {\small \tt garamlee@cryptolab.co.kr} \\
  {\small \tt \{minsoo9574, jhyunp, seungwonh, jhcheon\}@snu.ac.kr} \\
}

\begin{document}
\maketitle
\blfootnote{\textsuperscript{*}Equal contribution.}
\blfootnote{\textsuperscript{\textdagger}Corresponding author.}

\begin{abstract}
Embeddings, which compress information in raw text into semantics-preserving low-dimensional vectors, have been widely adopted for their efficacy.
However, recent research has shown that embeddings can potentially leak private information about sensitive attributes of the text,
and in some cases, can be inverted to recover the original input text.
To address these growing privacy challenges, we propose a privatization mechanism for embeddings based on homomorphic encryption, to prevent potential leakage of any piece of information in the process of text classification.
In particular, our method performs text classification on the encryption of embeddings from state-of-the-art models like BERT, supported by an efficient GPU implementation of CKKS encryption scheme.
We show that our method offers encrypted protection of BERT embeddings, while largely preserving their utility on downstream text classification tasks.
\end{abstract}
\section{Introduction}
In recent years, the increasingly wide adoption of vector-based representations of text such as
BERT, eLMo, and GPT~\citep{bert,elmo,gpt2},
has
called
attention to the privacy ramifications of embedding models.
For example, \citet{privacy_preserving_rep_adversarial,privacy_preserving_rep} show that sensitive information such as the authors' gender and age can be partially recovered from an embedded representation of text. 
\citet{infoleakage} report that 
BERT-based sentence embeddings can be inverted to recover up to 50\%–70\% of the input words. 

Previously proposed solutions such as $d_{\chi}$-privacy, a relaxed variant of local differential privacy based on perturbation/noise~\cite{privacybert},
require manually controlling the noise injected into embeddings,
to control the privacy-utility trade-off to a level suitable for each downstream task.
In this work, we propose a privacy solution based on \textit{Approximate Homomorphic Encryption}, which is able to achieve little to no accuracy loss of BERT embeddings on text classification\footnote{Code and data are available at: \url{https://www.github.com/mnskim/hebert}}, while ensuring a desired level of encrypted protection, i.e. 128-bit security.

Homomorphic Encryption (HE) is a cryptographic primitive that serves computations over encrypted data without any decryption process. 
While previous works have focused on homomorphic computation where the inputs are numerical data,
in applications such as privacy-preserving machine learning algorithms \cite{tmp_11}, logistic regression~\citep{tmp_1}, and neural network inference~\citep{tmp_4}, 
they have rarely been applied to \textit{unstructured} data such as text.
Recent works in this direction include
~\citet{word_embd}, who conduct sentiment classification over encrypted word embeddings using RNN. 
However, they use a simple embedding layer which maps words in a dictionary to real-valued vectors, and model training is only supported on plaintext. 
The most closely related work to ours is PrivFT~\citep{privft}, a homomorphic encryption based method for privacy preserving text classification built on fastText~\citep{fasttext}.

We next describe our approach, focusing on our distinctions from PrivFT:

\begin{itemize}[leftmargin=0.5em]
\item \textbf{BERT Embedding-based Method}: 
The principle behind PrivFT is to perform all neural network computations in encrypted state.
For this purpose, it adopts fastText~\cite{fasttext}, 
which takes bag-of-words vectors as input,
followed by a two-layer network and an embedding layer.
However, PrivFT does not utilize pre-training; 
as a consequence, the embedding matrix and classifer of PrivFT must be updated from scratch, 
taking several days to train on a single dataset.

We introduce a new method for text classification on encrypted data. 
The crux is to operate a simple downstream classifier on encryptions of semantically rich vector representations (i.e. BERT embeddings). 
By using rich input representations, our method significantly outperforms PrivFT, 
while the use of a simple downstream classifier on encrypted data makes our method much more practical. 
Importantly, by leveraging pretrained embeddings from models such as BERT, a state-of-the-art in many NLP tasks,
our method enables the training of a strong classifier in encrypted state within hours.
As such, our method is well positioned to take full advantage of
the recent trends in NLP, that rely on the language understanding capability of
increasingly larger pre-trained language models~\cite{gpt3,lm_scaling}.

\item \textbf{Better GPU Implementation}: 
As BERT representations are real-valued vectors, 
we adopt CKKS scheme, which is well-suited for dealing with real numbers compared to other HE schemes.
We develop an efficient GPU implementation of CKKS which greatly improves computation speed. 
While PrivFT also provides a GPU implementation of CKKS, their implementation lacks the \textit{bootstrapping} operation of CKKS. 
Inevitably, this limits the multiplicative depth of PrivFT, and it makes the method less scalable.
It also results in the use of less secure CKKS parameters which have roughly 80-bit security level. 
In contrast, our GPU implementation includes the bootstrapping operation, which allows unlimited number of multiplications. 
This enables us to use a higher degree polynomial approximation (which is key to achieving a high downstream accuracy), and more secure CKKS parameters (128-bit security level\footnote{An attacker needs $>2^{128}$ operations to recover the plaintext from a ciphertext with the current best algorithm.}). 
Moreover, with practicality in mind, we improved the implementation in terms of communication cost.
More precisely, we introduce a practical implementation of CKKS to significantly reduce the size of ciphertexts by more than $7.4\times$ compared to the rudimentary implementation.
\end{itemize}

We experimentally validate our approach on text classification datasets, 
showing that it offers encrypted protection of embedding vectors,
while maintaining utility competitive to plaintext on downstream classification tasks.
Additionally, we compare our method with PrivFT on homomorphic training on encrypted data, showing it outperforms PrivFT, with much improved training efficiency.

\section{Method}

\newcommand{\add}{\textsf{Add}}
\newcommand{\mult}{\textsf{Mult}}
\newcommand{\rot}{\textsf{Rotate}}
\newcommand{\bootstrap}{\textsf{Bootstrap}}
\newcommand{\ctxt}{\textsf{ct}}
\newcommand{\ptxt}{\textsf{m}}

\begin{figure}[t] 
\begin{center}

\includegraphics[width=1.0\linewidth]{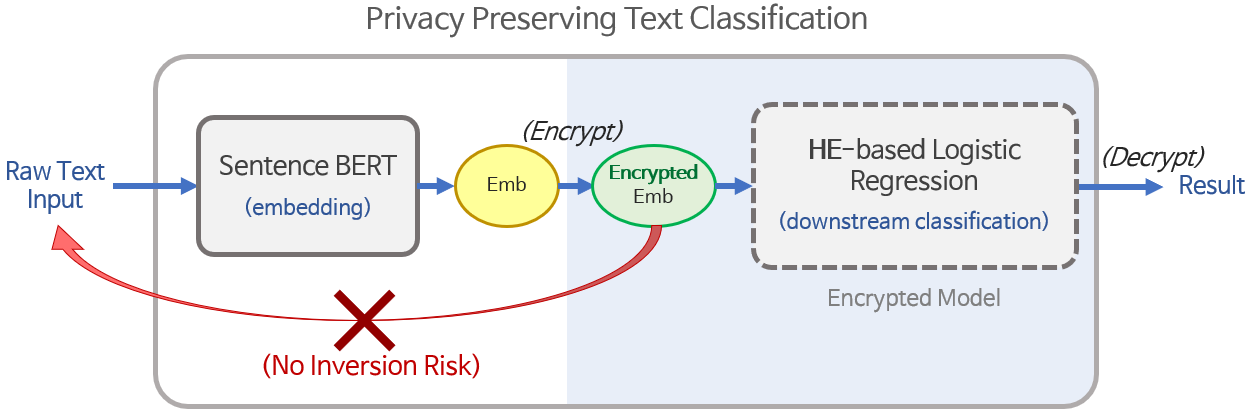}

\end{center}
\caption{\label{fig:Privacy Preserving Logistic Regression}\textbf{The Architecture of Text Classification.} The region shaded in light blue represents encrypted state. The privatization inference takes the following steps: 1) User generates sentence embedding. 2) User encrypts embedding. 3) Logistic regression in encrypted state is performed using encrypted embedding.}
\end{figure}

We focus on the scenario in which the user directly applies the privacy mechanism to the output embeddings from a neural text encoder, before passing it on to a service provider for usage in a downstream task.
Such a scenario is also referred to as a local privacy setting~\cite{privacybert}.
The privatization procedure $M_{priv}$ can be defined as follows:
\begin{align} \label{eq1}
    M_{priv}(x) = P(F_{emb}(x)) 
\end{align}
where $x$ is the raw text input, $F_{emb}$ is Sentence-BERT~\citep{sentencebert}\footnote{\url{https://www.sbert.net/}}, a popular pre-trained model for obtaining sentence embeddings, and $P$ denotes a privacy mechanism.
Next, we securely classify the text datum, $x$, by feeding its privatized embedding, $M_{priv}(x)$, to the downstream classification model. 
In this work, we adopt a logistic regression model (in encrypted state) as the downstream classifier.
Figure \ref{fig:Privacy Preserving Logistic Regression} demonstrates the entire privatization inference procedure,
starting with user's embedding of raw text and encryption of embedding, to the operation of the classifier in encrypted state and finally, the output of encrypted classification results. 
We note that the training process also can be performed in encrypted state as we describe in Section~\ref{sec:ours}.

\subsection{Baseline : $d_{\chi}$-privacy}
\label{sec:differential_privacy}
As a baseline, we implement $d_{\chi}$-privacy, a relaxation of noise-based local differential privacy (LDP).
\citet{privacybert} introduced $d_{\chi}$-privacy for single-token embeddings as privatization mechanism $P$. In the case of single-token embeddings, $d_{\chi}$-privacy can be achieved with respect to a chosen Euclidean distance by adding randomly sampled noise $N$ drawn from an $n$-dimensional distribution with density $p(N) \propto \exp(- \eta ||N||)$. 
That is, the privacy mechanism is $P(y) = y + N$, where $y$ denotes an embedding vector. 
In our work, we adopt the same mechanism to sentence embeddings. Following ~\citet{privacybert}, the noise $N \in \mathbb{R}^{n}$ is sampled as a pair $(r, p)$, where $r$ is the distance from the origin and $p$ is a point in $\mathbb{B}^n$ (the unit hypersphere in $\mathbb{R}^{n}$).
$r$ is sampled from Gamma distribution $\Gamma(n, \frac{1}{\eta})$ and $p$ is sampled uniformly over $\mathbb{B}^n$,
and $N$ is computed as $N=rp$.

\begin{table}[t]
\caption{\textbf{Size of Ciphertext/Training Time of Encrypted Logistic Regression.} We report the ciphertext size and training time for the ciphertext model. Note that plaintext and $d_{\chi}$-privacy classifiers have negligible training and inference times.}
\label{tab:size_time}
\vspace{-0.2cm}
\footnotesize
\centering
\resizebox{\columnwidth}{!}{
\begin{tabular}{l|ccc|ccc}
\toprule

& \multicolumn{3}{c|}{Twitter (2 classes)} & \multicolumn{3}{c}{SNIPS (7 classes; OvR)}\\

\cline{1-7} & \multicolumn{1}{c}{ciphertext} & \multicolumn{1}{c}{plaintext} & \multicolumn{1}{c|}{training time} &  \multicolumn{1}{c}{ciphertext} & \multicolumn{1}{c}{plaintext} & \multicolumn{1}{c}{training time} \Tstrut\Bstrut\\

\cline{2-7}
\multirow{1}{*}{train} & {\addstackgap{\shortstack[c]{1.4\\GB}}} & {\shortstack[c]{183.7\\MB}} &{\shortstack[c]{143.2\\sec/epoch}}   & {\shortstack[c]{11.4\\GB}} &{\shortstack[c]{206.5\\MB}} & {\shortstack[c]{1111.4\\ sec/epoch }}\\

\bottomrule
\end{tabular}
}
\end{table}

\subsection{HE Based Logistic Regression}\label{sec:ours}

We now describe our proposed privatization mechanism in detail. 
We adopt Eq.\ref{eq1},
with the privacy mechanism $P(y) = H(y)$, where $H$ is the homomorphic encryption. For downstream tasks, we feed the privatized embedding $M_{priv}(x)$ to an \textit{encrypted} logistic regression classifier.
By utilizing HE, an encrypted model and labels will be obtained after the training and inference process. Only the user who knows the secret key of HE can decrypt the results and get either the classifier or labels for the classification.

For the homomorphic encryption $H$, 
we adopt the CKKS scheme~\citep{tmp_6,tmp_7,tmp_8}. 
While the majority of HE schemes are being optimized for computations over finite fields, 
CKKS supports efficient computations over \textit{real numbers}, so it is advantageous in application to real world data. 
We refer the readers to the paper~\cite{tmp_6} for full details of CKKS.

CKKS is a levelled homomorphic encryption scheme, where the \textit{level} of each ciphertext indicates the
remaining number of times we can operate.\footnote{We remark that this \textit{level} is not related to the security level of CKKS. For example, in our implementation, all ciphertexts have $128$-bit security level regardless of the remaining number of operations.}
When we multiply two ciphertexts of level $l$, the output ciphertext has a level of $l-1$. 
Once the level of a ciphertext becomes too low, we can refresh its level higher by using the \textit{bootstrapping} technique  so that the number of possible operation times increases. 
For ciphertexts $\ctxt_1$ and $\ctxt_2$ of complex vector messages $\ptxt_1$ and $\ptxt_2$,
we summarize the operations of CKKS as follows:
\begin{itemize} 
\setlength{\itemindent}{-1em}
\setlength\itemsep{-0.3em}
    
    \item $\add(\ctxt_1,\ctxt_2)$: output a ciphertext  of $\ptxt_1+\ptxt_2$. 
    \item $\mult(\ctxt_1,\ctxt_2)$: output a ciphertext of $\ptxt_1\odot \ptxt_2$, where $\odot$ is the entry-wise multiplication.
    \item $\bootstrap(\ctxt_1)$:  output a ciphertext of $\ptxt_1$ at refreshed level. 
\end{itemize}
While it is prevalent to encrypt data into the top level, $L$, in this work, we encrypt the data into a lower level, $3$, to decrease the initial size of ciphertexts.\footnote{We use $L=29$ in our implementation.} Note that the ciphertexts are the privatized embeddings, so their size determines the communication cost. 
As shown in Table \ref{tab:size_time}, by using the lower level ciphertexts, we reduce the initial size of ciphertext by more than $7.4\times$ in both Twitter training dataset ($10.8$GB to $1.4$GB), and SNIPS training dataset ($85.3$GB to $11.4$GB).

Finally, we feed the output of our privatization mechanism to the next step, the training and inference of an encrypted logistic regression classifier. However, since CKKS supports only addition and multiplication while the logistic function $(1/(1+\text{exp}(-x))$ is a non-polynomial function, we evaluate the logistic function via its polynomial approximation. We use the minimax approximate polynomial ~\citep{tmp_10} of degree $15$ on $[-12, 12]$ that approximates the logistic function within the error of $0.00614$ on $[-12,12]$. 

\begin{table*}[t]
\caption{\textbf{Results of Logistic Regression Experiments.} For SNIPS, macro average of F1 over classes is reported and AUC denotes the average of of each class AUC. Bold and underline denote the ciphertext model and the noising model most comparable to it (measured by absolute difference of metric), respectively.
}
\label{tab:logreg_results}
\vspace{-0.2cm}
\footnotesize
\centering
\begin{tabular}{l|ccc|cc|cc|cc}
\toprule
\multirow{2}{*}{Model} &   \multicolumn{5}{c|}{Twitter}          &       \multicolumn{4}{c}{SNIPS}                \\

\cline{2-10} & \multicolumn{1}{c}{(Thresh)} & \multicolumn{1}{c}{Dev F1} & \multicolumn{1}{c|}{Test F1} & \multicolumn{1}{c}{Dev AUC}  & \multicolumn{1}{c|}{Test AUC}  &  \multicolumn{1}{c}{Dev F1} & \multicolumn{1}{c|}{Test F1} & \multicolumn{1}{c}{Dev AUC}  & \multicolumn{1}{c}{Test AUC} \Tstrut\Bstrut\\
\hline

\hline

Noising &&&&&&&&&\Tstrut\Bstrut\\
\cline{1-1}
\multicolumn{1}{l|}{$\eta$ = 50}     & 0.5149  & 0.3809 & 0.3337 & 0.7975 & 0.7991 & 0.7291 & 0.6944 & 0.9190 & 0.9106 \Tstrut\\
\multicolumn{1}{l|}{$\eta$ = 75}     & 0.5300  & 0.5226 & 0.4680 & 0.8847 & 0.8819 & 0.8818 & 0.8524 & 0.9689 & 0.9621 \\
\multicolumn{1}{l|}{$\eta$ = 100}    & 0.4997  & 0.5744 & 0.5323 & 0.9098 & 0.9105 & 0.9279 & 0.8990 & 0.9826 & 0.9776\\
\multicolumn{1}{l|}{$\eta$ = 125}    & 0.4555  & 0.6107 & 0.5760 & 0.9226 & 0.9245 & 0.9422 & 0.9190 & 0.9931 & 0.9844\\
\multicolumn{1}{l|}{$\eta$ = 150}    & 0.4843  & 0.6224 & 0.5939 &0.9234 &0.9332 & 0.9547 & 0.9303 & 0.9953 & 0.9879\\
\multicolumn{1}{l|}{$\eta$ = 175}    & 0.5128  & \underline{0.6404} & \underline{0.6065} &  \underline{0.9300} & \underline{0.9390} & \underline{0.9616} & \underline{0.9345} & \underline{0.9955} & \underline{0.9900}\Bstrut\\ 
\cline{1-10}
\textbf{Ciphertext}                  & 0.8635 & \textbf{0.6596} & \textbf{0.6361} & \textbf{0.9481} & \textbf{0.9535} & \textbf{0.9729} & \textbf{0.9402} & \textbf{0.9974} & \textbf{0.9948} \Tstrut\Bstrut\\
\cline{1-10}
Plaintext                 &0.4987   & 0.6625 & 0.6439 & 0.9536 & 0.9575 & 0.9787 & 0.9520 & 0.9977 & 0.9959\Tstrut\\
\bottomrule
\end{tabular}
\end{table*}
\subsection{Datasets}
To validate our approach in real-world scenarios, we conduct experiments on tasks with realistic privacy concerns and utility needs. We select text classification tasks on data in three settings:

\begin{itemize}

    \item \textbf{Tweets Hate Speech Detection}~\citep{twitter_dataset_source}\footnote{\url{https://huggingface.co/datasets/tweets_hate_speech_detection}}: Is a crowd-sourced dataset of Tweets for binary classification, where labels denote a Tweet as containing hate speech, if it has a racist or sexist sentiment associated with it. 
    We created random data splits for train/validation/test, with 11,634/3,197/4,795 examples, respectively.

  \item \textbf{SNIPS}~\citep{snips}: Is a dataset of crowd-sourced queries collected from the Snips Voice Platform, distributed among 7 user intents. It has been widely adopted in evaluating spoken language understanding (SLU) systems. We use the same data splits as \citet{snips_goo,snips_qin}, with 13,084/700/700 examples, respectively.
    
    \item \textbf{Youtube Spam Collection}~\citep{privft}\footnote{\url{https://archive.ics.uci.edu/ml/datasets/YouTube+Spam+Collection}}: 
    Is a public data set collected for spam research from UCI Machine Learning Repository, where five datasets are composed by 1,956 real messages extracted from five videos.  
    As train/validation/test splits are not provided, we created our own random splits, with 1,564/196/196 examples, respectively.
    
\end{itemize}
\section{Experiments}
\label{sec:experiments}

\subsection{Encrypted Sentence Classification}
\label{sec:experiments_classification}
Once sentence embeddings are extracted from Sentence-BERT for each input text, the vectors consist of 768 numerical values of 32-bit floating point from -1 to 1.
Then, a logistic regression model is trained for binary classification on the Twitter dataset, and multiclass classification on the SNIPS dataset, respectively.

To perform a fair comparison of the results of each approach, we keep the same implementation of logistic regression for plaintext,
as that of the ciphertext model.
Multiclass classification is performed as multiple separate binary logistic regression models for each class,
and we take the argmax from the combined results;  One-vs-Rest (OvR).
Experiments for noise-based $d_{\chi}$-privacy on plaintext are conducted in the same way,
using the privacy mechanism described in Section \ref{sec:differential_privacy}. 
Logistic regression parameters are optimized by SGD with Nesterov momentum. For all models, the best performing model and optimal threshold for F1 was identified by validation set performance.

For plaintext experiments, the following hyperparameters were used for training: Learning rate 3.0, gamma 0.9, batch size 256 for Twitter dataset, and learning rate 3.0, gamma 0.1, batch size 128 for SNIPS dataset. Both models were trained for 10 epochs.
For the parameters of the CKKS scheme, we selected the dimension $N=2^{17}$ and set the size of the maximum modulus $q_L$ to be $1540$ bits. 
We note that our CKKS parameters satisfy $128$-bit security level~\cite{lwe-estimator}.
For ciphertext experiments, the following hyperparameters were used for training:
Learning rate 3.0, gamma 0.9, batch size 512 for Twitter dataset, and learning rate 2.0, gamma 0.1, batch size 512 for SNIPS dataset. Both models were trained for 10 epochs.
Additionally, we developed an efficient parallelized CKKS implementation for bootstrapping with GPU acceleration for the encrypted logistic regression model.
For implementation, we use a dual-NVLink Nvidia Quadro RTX6000 GPU with 24 GiBs of memory,
on a server with a Intel Xeon Gold 6242R CPU (80 core)
and 125 GiBs of RAM.
\subsection{Embedding Inversion}
As a quantitative evaluation of inversion risk, we adopt sentence embedding inversion.
Introduced in \citet{infoleakage}, embedding inversion is an adversarial attack whose goal is to recover the original text (its tokens) from its embedding. In this work, we focus on \textit{black-box} inversion, 
where the adversary can only interact with the model by querying it to obtain embeddings, and is therefore more pertinent to real-world privacy considerations.

\section{Results}

\begin{table}[t]
\caption{\textbf{Sentence Embedding Inversion.} Black-box inversion of sentence embeddings on SNIPS. We report F1 for the task of recovering the input words from the sentence embedding. Ciphertext denoted in bold.}
\label{tab:sentence embedding inversion}
\vspace{-0.2cm}
\footnotesize
\centering
\begin{tabular}{l|ccc}
\toprule

Model & \multicolumn{1}{c}{(Thresh)} & \multicolumn{1}{c}{Dev F1} & \multicolumn{1}{c}{Test F1}  \\

\hline

Noising &&&\Tstrut\Bstrut\\
\cline{1-1}
\multicolumn{1}{l|}{$\eta$ = 50}     & 0.8  & 0.2082 & 0.1905 \Tstrut\\
\multicolumn{1}{l|}{$\eta$ = 75}     & 0.8  & 0.3078 & 0.2955  \\
\multicolumn{1}{l|}{$\eta$ = 100}    & 0.9  & 0.3587 & 0.3276 \\
\multicolumn{1}{l|}{$\eta$ = 125}    & 0.9  & 0.4164 & 0.3899 \\
\multicolumn{1}{l|}{$\eta$ = 150}    & 0.9  & 0.4572 & 0.4337 \\
\multicolumn{1}{l|}{$\eta$ = 175}    & 0.85  & 0.4919 & 0.4803 \Bstrut\\ 

\cline{1-4} 

\multicolumn{1}{l|}{\textbf{Ciphertext}}    & -  & \textbf{-} & \textbf{-} \Tstrut\\ 

\cline{1-4} 

\multicolumn{1}{l|}{Plaintext}    & 0.85  & 0.6705 & 0.6759 \Tstrut\\ 
\bottomrule
\end{tabular}

\end{table}

We report the results of our logistic regression experiments in Table \ref{tab:logreg_results}. 
We compare our approach, denoted as Ciphertext, 
with the Plaintext baseline, 
as well as the $d_{\chi}$-privacy from \citet{privacybert}, at different levels of the noise paramter $\eta$ (smaller indicates larger noise).
Measured by F1/AUC metric, our HE classifier achieves roughly $98.79\%/99.58\%$ and $98.76\%/99.89\%$ of the plaintext baseline classifier's performance on Twitter and SNIPS test sets, respectively,
indicating that our HE of embeddings is able to preserve their downstream utility to a significant degree.
We find that our model performs better at all noise levels considered in~\citet{privacybert} (up to $\eta=$ 175),
nearly matching the plaintext model.
On the other hand, for $d_{\chi}$-privacy, 
we observe a clear trade-off between increasing (via decreasing $\eta$) privacy protection
and classification performance.
As can be seen with $\eta=50,75$ on both datasets, the decrease in performance becomes greater as $\eta$ becomes smaller and privacy protection is prioritized.
Moreover, at any reasonable level of $\eta$, $d_{\chi}$-privacy cannot necessarily guarantee the complete elimination of inversion risk.

We next perform sentence embedding inversion experiments on SNIPS.
In Table \ref{tab:sentence embedding inversion}, we report the results 
at varying levels of $\eta$, using black-box inversion with a multi-label classification model as in~\citet{privacybert}. 
For plaintext, the degree of inversion risk is consistent with black-box inversion results from \citet{infoleakage},
who report F1 of 59.76 for inverting BERT-based sentence embeddings\footnote{Trained using Sentence-BERT objective on BookCorpus and Wikipedia data.},
indicating a high degree of invertibility. Our results show that, in order to significantly reduce inversion risk,
$d_{\chi}$-privacy requires low $\eta$ settings,
sacrificing downstream utility. In contrast, our method eliminates conventional risk of black-box inversion: Because all results of HE inference remain encrypted, and cannot be revealed without decryption with the user's secret key,
black-box inversion cannot be applied.
Therefore, 128-bit security level of homomorphic encryption guarantees practically complete protection from inversion, while offering significantly improved performance.

Finally, to directly compare our model with PrivFT, we conduct an experiment on the YTSC dataset, following the methodology in Section \ref{sec:experiments}. We report the results in Table \ref{tab:comparision to PrivFT}, along with PrivFT results on the same dataset from~\citet{privft}. We measure the test accuracy of the classifier, as well as the wallclock time required to perform encrypted training. 
We find that our method requires only $460.81$ seconds with a single GPU to achieve $90.8\%$ test accuracy, whereas PrivFT needs $5.04$ days with $8$ GPUs to obtain $86.3\%$ test accuracy. 
This amounts to roughly $\times$9,450 faster training per epoch, while achieving higher accuracy and utilizing 1/8\textsuperscript{th} the number of GPUs. These results experimentally 
validate our expectation that homomorphic encryption of pretrained embeddings significantly improves performance and efficiency.
\section{Conclusion}
\begin{table}[t]
\caption{\textbf{Comparison to PrivFT.} We report wallclock training time and test accuracy of binary spam classification. Ciphertext model results are denoted in bold.}

\label{tab:comparision to PrivFT}
\vspace{-0.2cm}
\footnotesize
\centering
\resizebox{\columnwidth}{!}{
\begin{tabular}{l|ccc}
\toprule

Model & \multicolumn{1}{c}{PrivFT} & \multicolumn{1}{c}{Ciphertext} & \multicolumn{1}{c}{Plaintext}   \\
\hline

\multicolumn{1}{l|}{(Num. GPUs)} &   8 & 1 & - \\
\multicolumn{1}{l|}{Training time} & 60.48 hrs/epoch  & \textbf{23.04 sec/epoch} & - \\
\hline
\multicolumn{1}{l|}{(Thresh)} &   - & 0.53 & 0.51 
\\
\multicolumn{1}{l|}{Test accuracy}    & 0.863  & \textbf{0.908}& 0.913
\\
\bottomrule 
\end{tabular}}
\end{table}

We propose a privatization mechanism based on homomorphic encryption which, by leveraging BERT pre-trained embeddings, enables efficient training of an HE logistic regression classifier with little to no loss of downstream utility. While our method compares favorably to $d_{\chi}$-privacy and PrivFT, 
we also note that there are some limitations. 
Since HE based models require higher computation costs compared to plaintext models, 
the challenge remains to adopt more complex models, such as neural networks, as downstream classifiers. Nevertheless, the privacy benefits and efficiency of our method makes it a suitable candidate for scenarios with real-world privacy concerns.

\section*{Acknowledgements}
We would like to show our gratitude to Younggi Lee and Seewoo Lee from CryptoLab for assistance with the experiments. We also greatly thank Jeonghwan Kim and Sungwoo Oh from KB Kookmin Bank who provided the initial research topic and suggestions for feasiblity in the real-world applications.

Cheon's team was supported by Institute of Information \& communications Technology Planning \& Evaluation~(IITP) grant funded by the Korea government (MSIT)~[NO.2020-0-00840, Development and Library Implementation of Fully Homomorphic Machine Learning Algorithms supporting Neural Network Learning over Encrypted Data, 50\%]. 
Hwang's team was supported by Microsoft Research Asia and IITP [(2022-00155958, High Potential Individuals Global Training Program) and (NO.2021-0-01343, Artificial Intelligence Graduate School Program (Seoul National University), 50\%].

\bibliography{anthology,custom}
\bibliographystyle{acl_natbib}

\end{document}